\documentclass[11pt, oneside]{article}    

\usepackage[caption=false,font=footnotesize]{subfig}
\usepackage{geometry}
\usepackage{graphicx}
\usepackage{color}
\usepackage{xcolor}
\usepackage{amssymb}
\usepackage{hyperref}
\usepackage[utf8]{inputenc}
\usepackage[T1]{fontenc}
\usepackage{url}
\usepackage{amsmath}
\usepackage{booktabs}
\usepackage{longtable}
\usepackage{amsfonts}
\usepackage{nicefrac}
\usepackage{microtype}
\usepackage{todonotes}
\usepackage{natbib}
\usepackage{titling}

\thanksmarkseries{arabic}

\newcommand{\mc}[1]{\ensuremath{\mathcal{#1}}}

\newcommand{\br}[0]{\\\vspace{-6px}\\}

\ifodd 1 \newcommand{\red}[1]{{\color{red}#1}} \else \newcommand{\red}[1]{} \fi

\title{
\vspace{-0.5em}
\textbf{Hide-and-Seek Privacy Challenge} \\
Synthetic Data Generation \textit{~vs.} Patient Re-identification \\
with Clinical Time-series Data
\vspace{-1em}
}
\author{
\normalsize James Jordon\thanks{
University of Oxford,
$^2$University of Cambridge,
$^3$University of California, Los Angeles,
$^4$University of Waterloo,
$^5$Amsterdam UMC,
$^6$Cambridge University Hospitals NHS Foundation Trust,
$^7$Microsoft Research,
$^8$Alan Turing Institute.
}
~~~~
\normalsize Daniel Jarrett$^2$                     ~~~
\normalsize Jinsung Yoon$^3$                       ~~~
\normalsize Tavian Barnes$^4$                      ~~~
\normalsize Paul Elbers$^5$                        \and
\normalsize Patrick Thoral$^5$                     ~~~
\normalsize Ari Ercole$^{2,6}$                     ~~~
\normalsize Cheng Zhang$^7$                        ~~~
\normalsize Danielle Belgrave$^7$                  \and
\normalsize Mihaela van der Schaar$^{2,3,8}$       ~~~
\\{\tt \normalsize  james.jordon@wolfson.ox.ac.uk}
}

\begin{document}
\maketitle

\begin{abstract}
The \textit{clinical time-series} setting poses a unique combination of challenges to data modeling and sharing. Due to the high dimensionality of clinical time series, adequate de-identification to preserve privacy while retaining data utility is difficult to achieve using common de-identification techniques. An innovative approach to this problem is \textit{synthetic data generation}. From a technical perspective, a good generative model for time-series data should preserve temporal dynamics, in the sense that new sequences respect the original relationships between high-dimensional variables across time. From the privacy perspective, the model should prevent \textit{patient re-identification} by limiting vulnerability to membership inference attacks. The NeurIPS 2020 Hide-and-Seek Privacy Challenge is a novel two-tracked competition to simultaneously accelerate progress in tackling both problems. In our head-to-head format, participants in the synthetic data generation track (i.e. ``hiders'') and the patient re-identification track (i.e. ``seekers'') are directly pitted against each other by way of a new, high-quality intensive care time-series dataset: the AmsterdamUMCdb dataset. Ultimately, we seek to advance generative techniques for dense and high-dimensional temporal data streams that are (1) clinically meaningful in terms of fidelity and predictivity, as well as (2) capable of minimizing membership privacy risks in terms of the concrete notion of patient re-identification.

\end{abstract}

\subsection*{Keywords}

Clinical Time-series Data; Data Privacy; Synthetic Data Generation; Patient Re-identification; Membership Inference Attack.

\subsection*{Competition Type}

Regular\vspace{0.5em}

\section{Competition Description}

\subsection{Background}\label{sec:background}

Coupled with advances in machine learning, the vast quantities of clinical data now stored in machine-readable form have the potential to revolutionize healthcare. At the same time, this enterprise is threatened by the fact that patient data are inherently highly sensitive, and privacy concerns have recently been thrown into sharp relief by several high-profile data breaches that have greatly undermined public confidence (see e.g. \cite{shah2017, price2019}). We seek novel methods capable of bridging the gap between data-hungry techniques in machine learning and privacy-conscious applications in healthcare settings.
\br
\textbf{Clinical Time-series}. Central among this balancing act is the development of techniques for modeling and sharing synthetic patient records in lieu of real data. However, the setting of clinical time-series data poses a unique combination of challenges to data modeling and sharing. Now, from a technical perspective, the learning problem in question is one of \textit{synthetic data generation}\textemdash a good generative model for time-series data should preserve temporal dynamics, in the sense that new sequences respect the original relationships between high-dimensional variables across time. Simultaneously from a social perspective, the privacy problem in question is one of \textit{patient re-identification}\textemdash a good generation technique should have the effect of preserving membership privacy, in the sense that the algorithm limits vulnerability of individual training instances to the risk of membership inference attacks.
\br
\textbf{(1) Time-series Generation}. Purely from the standpoint of generative modeling, the \textit{sequential} setting of clinical time-series data already presents a distinctive learning challenge. A model is not only tasked with capturing the distributions of patient features at each point in time, but it should also adequately reflect the potentially complex evolution of those variables over time. Existing methods directly apply the generative adversarial network (GAN) framework to temporal data, primarily by instantiating recurrent neural network (RNN) models as generators and discriminators (e.g. \cite{rcgan, c-rnn-gan, tcgan}). Such straightforward approaches neglect to leverage the autoregressive prior, and have been shown insufficient for ensuring that the network dynamics efficiently capture stepwise dependencies in the original training data (\cite{tgan}).
\br
\textbf{(2) Privacy and Identification}. Perhaps more importantly, the question of \mbox{synthetic data} generation cannot be divorced from concerns of \textit{privacy}. While de-identified data are commonly used for model development, existing notions of anonymity are often limited in scope: $k$-anonymity, $l$-diversity, and $\tau$-closeness are only aimed at protecting ``sensitive'' data (e.g. diagnoses) from an attack on a small number of quasi-identifiers (see \cite{sweeny2002, machanavajjhala2007} and \cite{Venkatasubramanian2007} respectively), and differential privacy (see \cite{dpbook}) has associated parameters that are difficult to interpret in terms of well-understood notions of leakage\textemdash such as vulnerability to membership inference attacks. In practice, the risk of patient re-identification is a pressing concern: Consider a rogue insurance company discriminating against high-risk patients per financial incentive. Such concerns caution medical institutions against releasing data for public research, hampering progress in the validation of novel computational models for real-world clinical applications.
\br
\textbf{Hide-and-Seek Challenge}. The NeurIPS 2020 Hide-and-Seek Privacy Challenge is a novel \textit{two-tracked} competition aiming to simultaneously accelerate progress in tackling both problems. In our head-to-head format, participants in the synthetic data generation track (i.e. ``hiders'') and the patient re-identification track (i.e. ``seekers'') are directly pitted against each other: the latter submit methods for launching membership inference attacks, while the former submit methods for synthesizing patient data that are robust to such attacks\textemdash all while maintaining faithfulness to the original data. Importantly, rather than falling back on fixed theoretical notions of anonymity, we allow participants on both sides to uncover the best approaches \textit{in practice} for launching or defending against privacy attacks. Ultimately, we seek to advance generative techniques for dense, high-dimensional temporal data streams that are (1) clinically meaningful in terms of fidelity and predictivity, as well as (2) capable of minimizing privacy risks in terms of the concrete notion of patient re-identification.

\begin{figure}[h]
\centering
\includegraphics[width=0.7\linewidth]{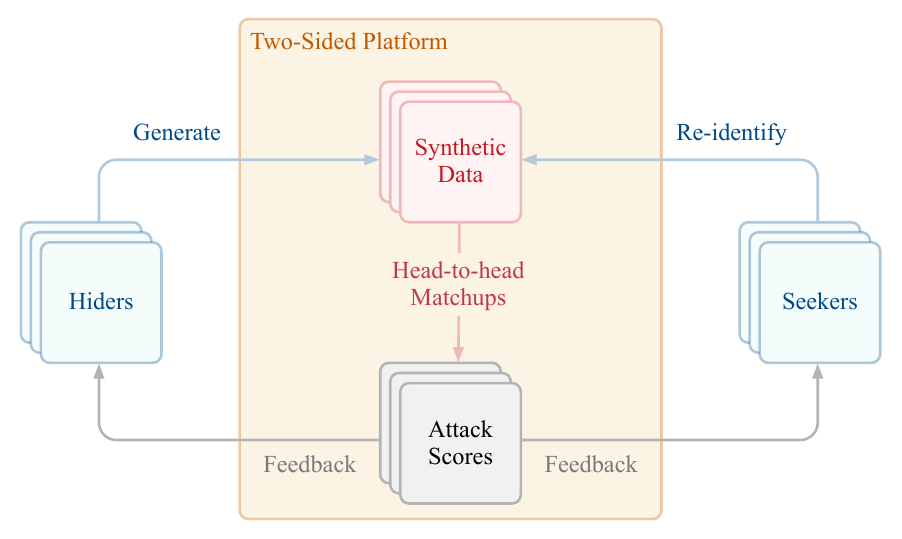}
\caption{Two-Tracked Competition. Head-to-head matchups between hiders (generating data) and seekers (re-identifying patients) will be conducted through the two-sided platform. (Not shown: Note that the synthetic data is separately evaluated on the hiders' side for faithfulness to the original data. See subsequent sections for scoring and ranking details.)}
\label{fig:pull}
\end{figure}

\subsection{Two-Track Format}

\begin{figure*}
\centering
\caption{Schematics and descriptions for the mechanics of submissions and evaluations.}
\label{fig:flow}
\subfloat[\textbf{Participants and Submissions}. Each generation-track participant $j\in\{1,...,N_{G}\}$ receives an initial $\mc{D}_{pub}$ for algorithm development; they each submit $\mc{M}^{(j)}_{G}$. Each inference-track participant $k\in\{1,...,N_{I}\}$ similarly receives the initial $\mc{D}_{pub}$ for algorithm development, and moreover has access to all $\mc{M}^{(j)}_{G},j\in\{1,...,N_{G}\}$; they each submit $\mc{M}^{(k)}_{I}$.]{
\hspace{-1.5em}
\includegraphics[width=0.52\linewidth]{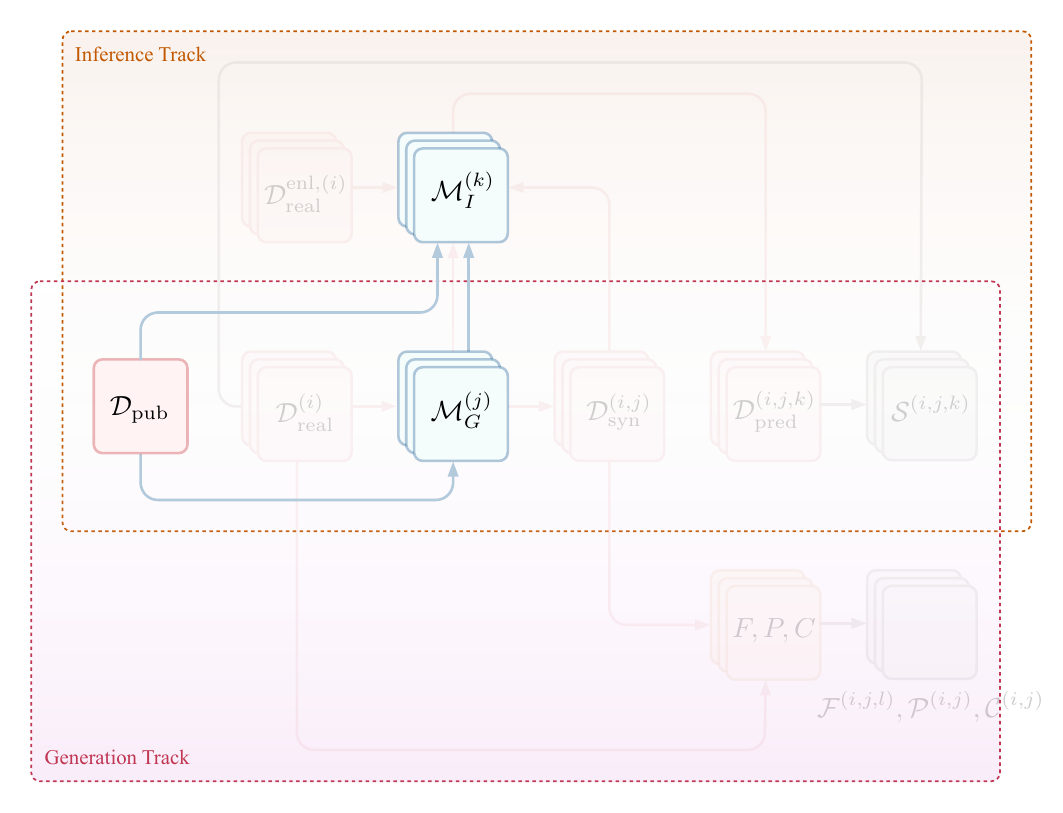}
\hspace{-1.5em}
}\hfill
\subfloat[\textbf{Generation Track Evaluation}. We sample $N$ subsets $\mc{D}_{\text{real}}^{(i)}$ from $\mathcal{D}_{priv}$, $i\in\{1,...,N\}$, each of which is used to train $\mc{M}^{(j)}_G$, outputting $\mc{D}^{(i, j)}_{\text{syn}}$. Each $\mc{D}_{\text{real}}^{(i)}$ is also randomly enlarged to form $\mc{D}_{\text{real}}^{\text{enl},(i)}$. Submissions qualify if the degradation from TRTR to TSTR satisfies Equations 4--6. Once qualified, the overall score for each $j$ is given by Equation 7.]{
\hspace{-1.5em}
\includegraphics[width=0.52\linewidth]{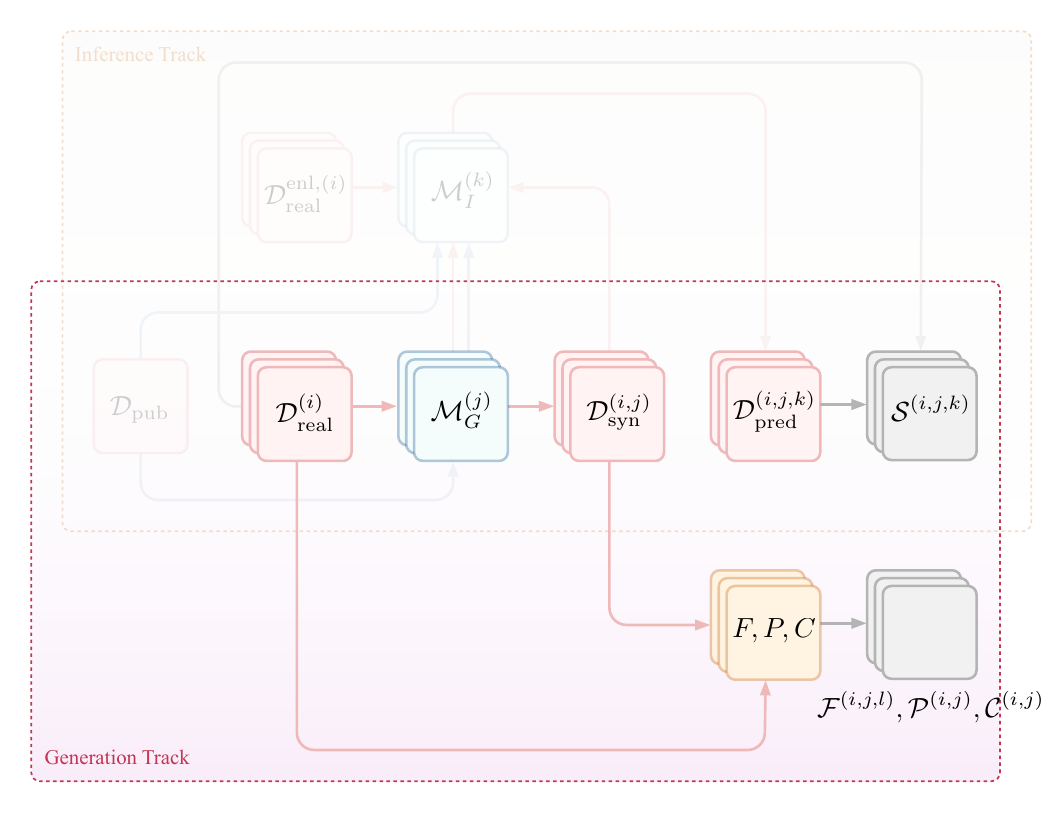}
\hspace{-1.5em}
}

\subfloat[\textbf{Inference Track Evaluation}. We run each inference algorithm $\mc{M}^{(k)}_I$ on the tuple $(\mc{M}^{(j)}_G, \mc{D}_{\text{real}}^{\text{enl},(i)}, \mc{D}^{(i, j)}_{\text{syn}})$, which outputs predictions $\mc{D}_{\text{pred}}^{(i,j,k)}$ that aim to approximate $\mc{D}_{\text{real}}^{(i)}$. The score for each $k$ is given by Equation 3.]{
\hspace{-1.5em}
\includegraphics[width=0.52\linewidth]{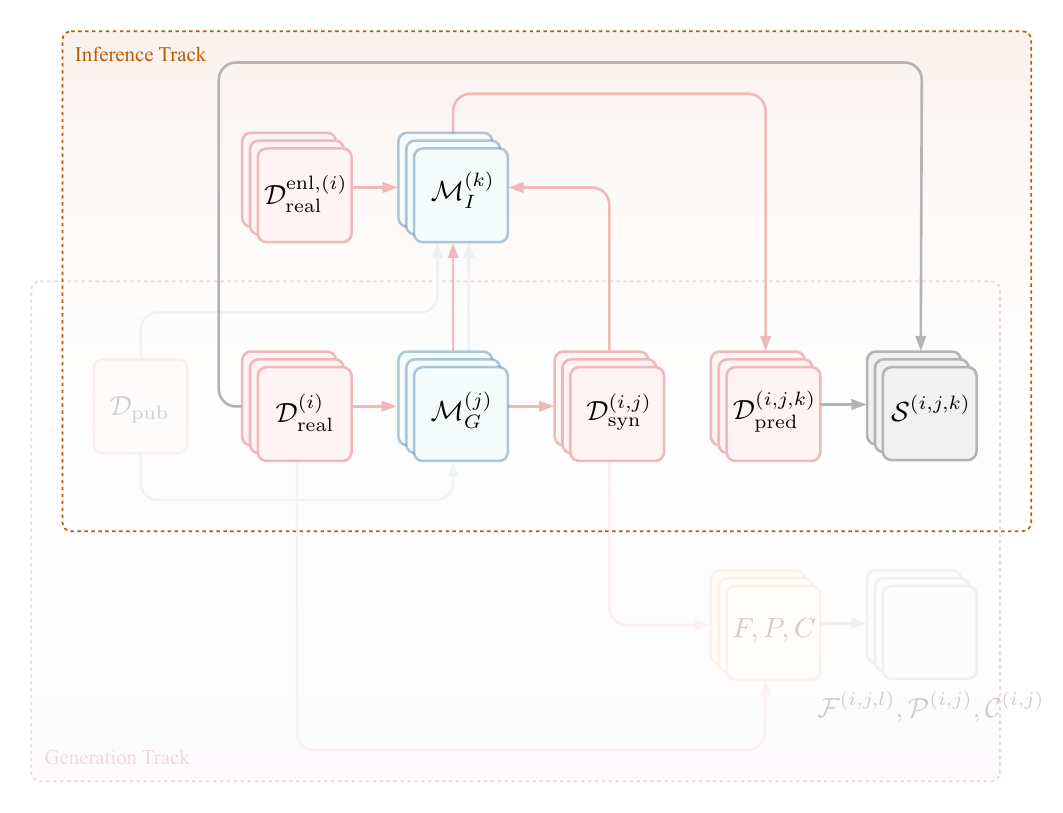}
\hspace{-1.5em}
}\hfill
\subfloat[\textbf{Complete Competition Setup}. Blue lines denote the tasks of participants in submitting algorithms (blue boxes). Red lines denote the flow of input and output datasets (red boxes) to and from algorithms. Gray lines denote the computation of performance scores (gray boxes). The orange box denotes a collection of standard prediction tasks.]{
\hspace{-1.5em}
\includegraphics[width=0.52\linewidth]{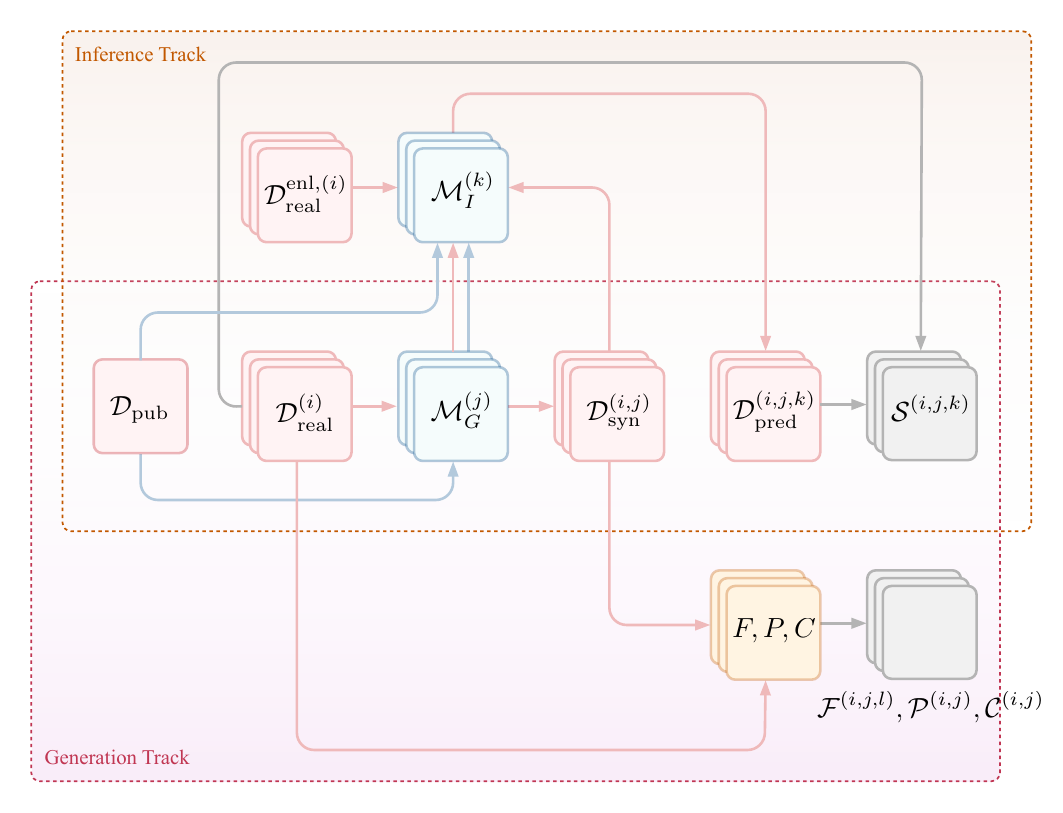}
\hspace{-1.5em}
}
\end{figure*}

This competition provides a \textit{two-sided platform} for synthetic data generation and patient re-identification methods to compete among and against each other. Our aim is to understand\textemdash through the practical task of membership inference attacks\textemdash the strengths and weaknesses of machine learning techniques on both sides of the privacy battle, in particular to organically uncover what existing (and potentially novel) notions of privacy and anonymity end up being the most meaningful in practice. We therefore invite participants to compete in either or both of two submission tracks of the interactive challenge: (1) the \textit{hider} (i.e. synthetic data generation) track, and (2) the \textit{seeker} (i.e. patient re-identification) track.
\br
\textbf{Structure}. Over the course of 3 months, models from both tracks will be pitted against each other in head-to-head matchups to determine (1) which seekers can re-identify the most patients, and (2) which hiders can best generate synthetic data resilient against privacy attacks (while retaining the statistical properties of the original data streams). The evolving marketplace of submissions on both sides will present live challenges: as more robust generative methods are submitted, re-identification methods will require ongoing adaptation to remain competitive; likewise, as stronger re-identification methods are submitted, generative methods will need to better protect against membership inference attacks. See Figure \ref{fig:pull} for a high-level overview, and Section \ref{sec:setup_detail} for a detailed discussion of the competition structure.
\br
\textbf{Evaluation}. Each head-to-head matchup is a zero-sum game. Seekers will be scored according to their accuracy on the membership inference task over each hider submission\textemdash that is, in correctly identifying whether a given instance was employed in the process of generating a given synthetic dataset. In the opposite direction, hiders will be scored according to how well their generation algorithms hold up to membership inference attacks. In addition, hider submissions are required to adequately capture the feature and temporal correlations in the original data; accordingly, they must also first pass a minimum-quality bar (in terms of fidelity and predictivity) in order to qualify for competition. (Although the trade-off between quality and privacy is {\em very} interesting in its own right, for the purpose of fair comparison we fix the former to allow ranking in terms of the latter.) See Section \ref{sec:scoring_detail} and Figure \ref{fig:flow} for a detailed discussion and schematic of the evaluation and scoring procedure.
\br
\textbf{Dataset}. This challenge will be held via clinical data from the intensive care unit (ICU). ICU admissions represent some of the most data-dense patient episodes in healthcare, and these patients represent some of the sickest. Unlike other healthcare domains, ICU data is characterized by its granular, sequential nature, its high-dimensionality and variety of data types, as well as heterogeneous sampling patterns and frequencies. This combination of challenges poses distinctive complexities for modeling; at the same time, it offers huge potential for improving patient care in real-time settings of life-and-death decision-making\textemdash where patients are often at risk of deterioration over the span of hours or minutes. Crucially, while a range of diverse models have been investigated in medical literature (see e.g. reviews of \cite{shillan2019, fleuren2020}), they are largely based on a small number of publicly available datasets. To date, the availability of alternative high-quality, dense, and high-dimensional datasets for verifying model generalizability has been limited\textemdash precisely due to concerns of privacy. See Section \ref{sec:dataset} for details of the new dataset we introduce.

\subsection{Novelty and Impact}

\subsubsection{Healthcare Novelty and Impact}\label{sec:healthcare_novelty}
Success in this competition will make headway in bridging the gap between machine learning and real-world healthcare applications. On the one hand, privacy is of paramount importance if medical data is to be disseminated for public research. On the other hand, it is critical that clinical applications be thoroughly tested in a wide variety of settings prior to deployment:
\br
\textbf{Generalizability} is of particular concern, since remarkable variations exist across different populations on the level of demographics, institutions, regions, and healthcare delivery systems. Consider ICU data: Without doubt, several publicly available datasets such as the Medical Information Mart for Intensive Care (MIMIC) (\cite{mimic}) have been instrumental in driving technical progress in statistical and machine learning applications. However there are well-known differences in intensive care delivery models globally (for instance, due to the greater availability of beds, US intensive care patients tend to have far lower acuity than those in Europe). Models developed and tested on US data simply do not generalize to European settings (see e.g. discussions in \cite{ercole2017}). To rigorously develop and validate models for generalizability, much more diverse datasets need to be made available if healthcare is to safely benefit from advances in machine learning.
\br
\textbf{Privacy}. The global legal framework protecting sensitive data (such as healthcare data) is a hodge-podge of ideas. For example, the US Health Insurance Portability and Accountability Act (HIPAA) and the corresponding European General Data Protection Regulations (GDPR) impose differing\textemdash and often ambiguous\textemdash constraints to limit data sharing. Both are deliberately vague in specifications of anonymity, relying only on intuitive notions of ``low probability of re-identification''. However neither ``low probability'' nor ``re-identification'' are well defined concepts. In the case of the latter, the concept of ``re-identification''' has typically been interpreted through notions of $k$-anonymity (\cite{sweeny2002}), $l$-diversity (\cite{machanavajjhala2007}),  $\tau$-closeness (\cite{Venkatasubramanian2007}), etc. As noted above, these only aim to protect ``sensitive'' data elements from attacks on a small tuple of quasi-identifiers.
\br
Novel methodologies are urgently needed to overcome current \textit{privacy} concerns that prevent valuable datasets from accelerating technical research to improve \textit{generalizability} for real-world clinical applications. Methods for generating high quality synthetic clinical time-series data robust to privacy attacks have the potential to be game-changing in the clinical arena.

\subsubsection{Technical Novelty and Impact}\label{sec:technical_novelty}
The competition is challenging and highly relevant to multiple machine learning communities; the ``hide-and-seek''' setup easily allows enthusiasts to tackle two different sets of challenges:
\br
\textbf{Hider Challenge}. In the ``hider'' challenge, a good generative model is needed for the synthetic data generation task\textemdash one that accounts for the distributions and correlations among and across both static and temporal features. Moreover, to be able to ``hide'' from a ``seeker'', various existing or novel concepts pertaining to anonymity and privacy may be considered in terms of their practical effectiveness. Joining this track of the competition is without doubt of interest to a large fraction of the machine learning community (e.g. any student or enthusiast with working knowledge of GANs may readily attempt the task).
\br
\textbf{Seeker Challenge}. In the ``seeker'' challenge, we offer a novel setup to experiment with developing and performing attacks on machine learning models. The privacy attack is rooted in real-world data privacy needs (esp. in healthcare), and has very different properties compared with gradient-based adversarial attacks. Attacks can be either white-box or black-box. This will raise new challenges to both the security and adversarial attack communities.
\br
\textbf{An Interdisciplinary Problem}. Owing to the rich structure of the data, both sides of the competition will benefit from research and methods inspired by self-supervised learning, automatic machine learning, etc. for model design. In addition, there is an implicit requirement for higher-level notions of generalization as well: As the competition progresses, more privacy attacking methods and more generative models will surface, and methods developed by participants will need to generalize to an increasingly wide range of opponent models.
\br
\textbf{Related Competition}. We are aware of the following related competition: \textit{NIST 2018 Differential Privacy Synthetic Data Challenge}. This challenge tasked participants with creating new methods, or improving existing methods of data de-identification, while preserving the dataset's utility for analysis. Competitors participated in three marathon matches on the Topcoder platform with the goal of designing, implementing, and proving that their synthetic data generation algorithm satisfied differential privacy. All solutions were required to satisfy the differential privacy guarantee, a provable guarantee of individual privacy.
\br
Our competition differs significantly from this. Rather than requiring that a model satisfy some provable theoretical notion of privacy (such as differential privacy), the privacy of an algorithm will be defined in terms of its \textit{practical} susceptibility to membership inference attacks on the platform. Our competition pits generators (`hiders') and re-identifiers (`seekers') against each other. While provable notions of privacy might be used by teams submitting to our generation track, the head-to-head format of our competition will enable practical evaluation of such notions, furthering the insights found in \citet{midpcomp}.\\
\br
\textbf{Methodological Challenges}. In terms of time-series data generation, little work has been done in the realm of private generation of {\em temporal} data \citep{rcgan}. However, many private data generation methods focusing on static data have been proposed \citep{pategan, dpgan, uclanesl_dp_wgan, adsgan}. The goal of this competition is to bring focus to the more difficult task of temporal data, which is clearly more rewarding and applicable to real-world applications, healthcare and otherwise.

\subsubsection{Dataset Novelty}
\label{sec:dataset}

This challenge will use a new dataset. AmsterdamUMCdb was developed and released by Amsterdam UMC in the Netherlands and the European Society of Intensive Care Medicine (ESICM). It is the first freely accessible comprehensive and high resolution European intensive care database. It is also first to have addressed compliance with General Data Protection Regulation (GDPR, EU 2016/679) using an extensive risk-based de-identification approach. However, both ESICM and Amsterdam UMC aim to continuously evaluate and if necessary improve privacy while maintaining usability.
\br
AmsterdamUMCdb contains real data from critically ill patients from a mixed surgical-medical tertiary referral centre for intensive care medicine with up to 32 intensive care beds and up to 10 high dependency beds. It was released in early 2020 as comma separated value files, with a total uncompressed size of approximately 78 GB. Access may be requested through \url{https://www.amsterdammedicaldatascience.nl} and downloaded from \url{https://doi.org/10.17026/dans-22u-f8vd}. Detailed descriptions of the data schema and sample code to interact with the data are available on the AmsterdamUMCdb GitHub repository at \url{https://github.com/AmsterdamUMC/AmsterdamUMCdb}. 
\br
AmsterdamUMCdb contains approximately 1 billion clinical data points related to 23,106 admissions of 20,109 unique patients between 2003 and 2016. The released data points include patient monitor and life support device data, laboratory measurements, clinical observations and scores, medical procedures and tasks, medication, fluid balance, diagnosis groups and clinical patient outcomes. Data granularity depends on the type of data and admission year, but is up to 1 value every minute for data from patient monitor and life support devices. The data is much richer and granular than those in other well known freely available intensive care databases, such as MIMIC and is comprised of patients with higher illness acuity than is found in US datasets.

\section{Tasks and Evaluation}
The following outlines the tasks and how they will be evaluated. Specific details of various parameters (e.g. number of experiments per evaluation $K$, fidelity cutoff $f$) outlined here will be released with the documentation in our starter kit.

\subsection{Challenge Tasks}\label{sec:setup_detail}

The competition will be a continuous process taking place over 3 months, during which participants are free to submit entries (i.e. algorithms) to either side of the competition. A complete evaluation procedure of all submitted entries will be conducted on a monthly basis, from which a leaderboard will be constructed for each track\textemdash respectively ranking submissions in order of performance.
\br
\textbf{Task for Hiders}
\br
In the \textit{synthetic data generation} track, participants are tasked with developing an algorithm that generates synthetic data on the basis of real data. Their submission must be an algorithm $\mc{M}_G$ (i.e. not just a trained model), whose {\bf input} will be random subsets of an unseen subset of the dataset, 
$$\mc{D}_{real} \subset \mc{D}\,,$$
and whose {\bf output} is a synthetic dataset,
$$\mc{D}_{syn} = \mc{M}_G(\mc{D}_{real})\,,$$
that contains entries from the same space as entries in the original dataset. At competition launch, participants will be given a subset of the dataset, $\mc{D}_{pub} \subset \mc{D}$. They will be free to use this data to develop their algorithm and perform preliminary hyper-parameter selection but may not use it to pre-train/initialise a model's weights.
\br
The synthetic data generated by each model will be evaluated in two ways: (1) similarity to the real data; and (2) resistance to re-identification. For each model this will be done on $K$ random subsets of the non-public data, i.e. $\mc{D}_{real}^1, ..., \mc{D}_{real}^{K} \subset \mc{D}_{priv} =  \mc{D} \setminus \mc{D}_{pub}$.
\br
\textbf{Task for Seekers}
\br
In the \textit{patient re-identification} track, participants are tasked with developing an algorithm that performs membership inference (aka Patient Re-identification) on synthetic data generation algorithms. Their submission must be an algorithm, $\mc{M}_R$ (which may contain trained models from the public data), whose {\bf input} will be tuples of the form
\begin{equation}
    (\mc{M}_G, \mc{D}_{syn}, \mc{D}_{real}^{enl})
\end{equation}
where $\mc{M}_G$ is an indicator for the generation algorithm used, $\mc{D}_{syn}$ is syntheic data generated by $\mc{M}_G$ and $\mc{D}_{real}^{enl}$ is a random subset of $\mc{D}$ that contains the real data, $\mc{D}_{real}$, used to generate $\mc{D}_{syn} = \mc{M}_G(\mc{D}_{real})$ (i.e. a randomly {\em enlarged} copy of $\mc{D}_{real}$). The {\bf output} must be a classification of each element of $\mc{D}_{real}^{enl}$, in which the goal is to classify the elements as:
\begin{center}
`{\em in $\mc{D}_{real}$}' or `{\em not in $\mc{D}_{real}$}'
\end{center}
or equivalently the output must be a subset $\mc{D}_{pred} \subset \mc{D}_{real}^{enl}$ corresponding to the elements the algorithm classifies as being `{\em in $\mc{D}_{real}$}'.
\br
At competition launch, participants will be given the same public dataset as the generation track, $\mc{D}_{pub} \subset \mc{D}$. In addition, as each generation algorithm is submitted it will be made publicly available alongside 10 synthetic datasets generated using 10 random (known) subsets of the public data (so that re-identification track participants do not need to - but are still welcome to - run the generation models themselves).
\br
Re-identification algorithms will be evaluated on $K$ synthetic datasets generated by each generation algorithm according to their classification accuracy. The synthetic datasets used for evaluation will be generated on the basis of random subsets of the {\em unseen} data, $\mc{D}_{priv}$.

\subsection{Scoring and Ranking}\label{sec:scoring_detail}
Suppose we receive submissions $\mc{M}_G^1, ..., \mc{M}_G^{N_G}$ to the generation track and $\mc{M}_R^1, ..., \mc{M}_R^{N_R}$ to the re-identification track. Let $\mc{D}_{real}^i \subset {\mc{D}_{real}^{enl}}^i\subset \mc{D}_{priv}$ be $K$ random subsets (and enlarged subsets) of the private data. For each generation algorithm and each real dataset we generate a synthetic dataset $\mc{D}_{syn}^{i, j} = \mc{M}_G^{j}(\mc{D}_{real}^i)$.
\br
\textbf{Evaluating Seekers}
\br
For each $i = 1, ..., K$ and $j = 1, ..., N_G$, a re-identification algorithm, $\mc{M}^k_R$, is assigned a score $\mc{S}^{i, j, k} \in [0, 1]$ according to its classification accuracy, given by:
\begin{align}
    \mc{S}^{i, j, k} = \frac{|\mc{D}^{i, j, k}_{pred} \cap \mc{D}^i_{real}| + |{\mc{D}^{i, j, k}_{pred}}^c \cap {\mc{D}^i_{real}}^c|}{|{\mc{D}_{real}^{enl}}^i|}
\end{align}
where $^c$ denotes the compliment of the set (within ${\mc{D}_{real}^{enl}}^i$) and $\mc{D}^{i, j, k}_{pred} = \mc{M}_R^k(\mc{M}_G^j, \mc{D}_{syn}^{i, j}, {\mc{D}_{real}^{enl}}^i)$.
\br
To create an overall score for a re-identification algorithm, we will average the score of the algorithm across the $K$ synthetic datasets for {\em each} generation algorithm. The overall score, $\mc{S}_O$, of re-identification algorithm, $\mc{M}^k_R$, is given by
\begin{equation}
    \mc{S}^k_O = \frac{1}{K \times N_G} \sum_{i=1}^{N_G} \sum_{j=1}^{10} \mc{S}^{i, j, k} \,.
\end{equation}
\br
\textbf{Evaluating Hiders}
\br
Before being ranked according to robustness to re-identification, algorithms in the generation track will be required to meet a minimum quality bar in order to qualify. Each of the $K$ synthetic datasets associated with each generation algorithm will be evaluated on their similarity to the real data. We will measure the quality of each synthetic dataset by training models on the synthetic data and then testing them on the real data. This will be compared to the performance of the same algorithms trained and tested on the real data. For this we consider 3 tasks that the synthetic data must be suitable for: (1) feature prediction, (2) sequential prediction, and (3) time-series classification.
\\ \\
{\bf (1)} For the first task we will randomly select $N_f$ features. For each of these features, we will train a predictive model, $F^{i, j, l}, l=1, ..., N_f$, on each synthetic dataset, $\mc{D}_{syn}^{i, j}$. To do this, the remaining features (all but the one being predicted) will be used as input. The predictive model's performance (RMSE for continuous features, AUROC for binary, accuracy for categorical), will be evaluated in the real data $\mc{D}_{real}^{i}$, producing a feature prediction score for each synthetic dataset, $\mc{F}^{i, j, l}$.
\\ \\
{\bf (2)} For the second task, a 1-step-ahead prediction model, $P^{i, j}$, will be trained using each synthetic dataset, $\mc{D}_{syn}^{i, j}$. The prediction model will predict the feature values at time step $t+1$ based on the entire history up to and including time step $t$. The prediction models will be tested in the real data $\mc{D}_{real}^{i}$ that corresponds to the synthetic data, producing a sequential prediction score for each synthetic dataset, $\mc{P}^{i, j}$.
\\ \\
{\bf (3)} For the final task, the goal will be classification based on an input data-stream. A classification model, $C^{i, j}$ that predicts a label (such as mortality, respiratory failure, cardiovascular failure etc.) will be trained on each synthetic dataset, $\mc{D}_{syn}^{i, j}$, and then tested in the real dataset, $\mc{D}_{real}^{i}$, producing a classification score, $\mc{C}^{i, j}$.
\\ \\
A score for each of these tasks will similarly be computed by training and testing within the real datasets $\mc{D}_{real}^i$, producing scores $\mc{F}^{i, l}, \mc{P}^{i}, \mc{C}^{i}$, $l = 1, ..., 10$. A generation algorithm, $\mc{M}_G^j$, will qualify if and only if it satisfies all of the following for all $i = 1, ..., K$:
\begin{align}
    \mc{F}^{i, j, l} &> f\mc{F}^{i, l} \text{ for all $l = 1, ..., N_f$ } \\
    \mc{P}^{i, j} &> f\mc{P}^i \\
    \mc{C}^{i, j} &> f\mc{C}^i
\end{align}
where $f \in [0, 1]$ - i.e. the models trained on the synthetic data must achieve at least fraction $f$ of the performance of the models trained on the real data, for {\em every} task.
\br
Models that qualify will then be ranked according to their robustness to re-identification (the quality scores are no longer relevant once the bar has been passed). The re-identification score, $\mc{R}^{j}$, of generation algorithm $\mc{M}_G^j$ is given by
\begin{equation}
    \mc{R}^j = \max_k \sum_{i = 1}^{K} \mc{S}^{i, j, k}
\end{equation}
where a {\em lower} score is better. This score corresponds to how well the {\em best} performing re-identification algorithm (for the given generation model) is able to re-identify {\em on average} across the different synthetic datasets generated by $\mc{M}_G$.

\subsection{Baseline Algorithms}

\textbf{Hider Baselines}
\br
We will provide 2 baseline models for the synthetic data generation task: (1) a simple noise-addition baseline; and (2) TimeGAN \citep{tgan}. The algorithms will be made publicly available in the form of source code. These will primarily be of use to the re-identification track as a starting point for developing their algorithms.
\br
\textbf{Seeker Baselines}
\br
We will have two model-agnostic baselines available at launch: (1) a simple nearest neighbour method; (2) a classifier based method. Details of both will be given in our starter kit documentation.

\subsection{Tutorial and Documentation}
We will use the TimeGAN baseline to provide an example submission to the generation track. We will also provide an example submission to the re-identification track with placeholders to be replaced by the user.

\section{Organizational Aspects}
\subsection{Protocol}
\textbf{Submissions and Procedure}
\begin{enumerate}
\item \textit{Generation Track}: Each generation-track participant $j\in\{1,...,N_{G}\}$ receives the (same) initial set $\mc{D}_{G}$ of samples for algorithm development; each submits an algorithm $\mc{M}^{j}_{G}$.
\item \textit{Inference Track}: Each inference-track participant $k\in\{1,...,N_{R}\}$ receives the (same) $\mc{D}_{G}$, and also has access to algorithms $\mc{M}^{j}_{G},j\in\{1,...,N_{G}\}$; each submits an algorithm $\mc{M}^{k}_{I}$.
\item \textit{Submissions Platform}: 
\begin{enumerate}
    \item We sample $K$ random subsets ${{\mc{D}}_{real}^{enl}}^{i}$ of records from the privately held dataset $\mathcal{D}_{priv}$, $i\in\{1,...,K\}$; for each subset, we further randomly sample 50\% of these records as $\mc{D}_{\text{real}}^{i}$.
    \item We train each generation algorithm $\mc{M}^{j}_G$ on $\mc{D}_{\text{real}}^{i}$, outputting synthetic datasets $\mc{D}^{i, j}_{syn}$.
    \item We run each inference algorithm $\mc{M}^{k}_I$ on $(\mc{M}^{j}_G, \mc{D}^{i, j}_{\text{syn}}, {\mc{D}_{real}^{enl}}^{i})$, outputting predictions $\mc{D}_{pred}^{i}$.
    \item We compute quality scores, $\mc{F}^{i, j, l}, \mc{P}^{i, j}, \mc{C}^{i, j}$ for $i,l \in \{1, ..., K\}, j \in \{1, ..., N_G\}$.
    \item We compute scores $\mc{S}^{i, j, k}$ for each combination of $i\in\{1,...,K\}$, $j\in\{1,...,N_{G}\}$, and $k\in\{1,...,N_{R}\}$, using only the generation algorithms that exceed the required quality scores. Using these scores we compute $\mc{S}_O^k$ for $k \in \{1, ..., N_R\}$ and rank re-identification algorithms according to this. We also compute $\mc{R}^j$ for $j \in \{1, ..., N_G\}$ and rank {\em qualifying} generation algorithms according to this. All scores are shown to all participants.
\end{enumerate}
\end{enumerate}
~
\\
\vspace{-2em}
\\
CodaLab (https://competitions.codalab.org/competitions/) will be used to host the competition.

\subsection{Rules}
\begin{itemize}
    \item ML-AIM lab, Microsoft Research Cambridge and Amsterdam UMC employees can participate but are ineligible for prizes.
    \item Participants that have access to the AmsterdamUMCdb dataset will be required to declare this and will be ineligible for prizes.
    \item Any participants that are ineligible for prizes will not contribute to the scores of other teams.
    \item To be eligible for the final scoring, participants are required to release the code of their submissions as open source.
    \item If a submission does not run successfully it is the participants' responsibility to debug it. Participants will be allowed to attempt to submit at most once per day.
    \item Generation algorithms may only use the public data to define and tune hyper-parameters of their algorithm but may not use the public data to initialise/pre-train a model. Winning algorithms will be closely inspected to ensure compliance before prizes are awarded.
    \item Each generative algorithm will be required to run within a specific time on a given GPU. Details will be provided in starter kit documentation.
    \item Each re-identification algorithm will be required to run within a specific time on a given GPU. Details will be provided in starter kit documentation.
\end{itemize}

\subsection{Schedule}

\begin{itemize}
\item {\bf May 11, 2020.} Launch website with announcement and competition rules. Begin
advertising the competition.
\item {\bf July 1 - October 1, 2020.} Competition runs. Generation submissions will be publicly available once submitted.
\item {\bf October 1, 2020.} Deadline for the final submission.
\item {\bf October 1 - November 16, 2020.} Organizers evaluate submissions.
\item {\bf November 16, 2020.} Announce competition results and release evaluation data. Release code of all re-identification algorithms.
\end{itemize}

\section{Resources}

\subsection{Organizing Team}

\begin{longtable}{ p{.7\textwidth}  p{.3\textwidth} } 
& {\bf Roles} \\
\textbf{James Jordon} is an Engineering Science PhD student at the University of Oxford. His primary research focus has been on generative models and their use for various tasks such as synthetic data generation, treatment-effect estimation and feature selection. He has published papers in several leading machine learning conferences including NeurIPS, ICML and ICLR. & \textbullet Lead Coordinator \newline \textbullet Competition Design \newline \textbullet Evaluation Design \newline \textbullet Advertisement \newline \textbullet Evaluator \newline \textbullet Baseline method\newline provider\\ \\

\textbf{Daniel Jarrett} is a Mathematics PhD student at the University of Cambridge. His primary research focus has been on representation learning for predictive, generative, and decision-making problems over time with a focus on healthcare. He has published in various journals and conferences including ICLR, NeurIPS, AISTATS, and The British Journal of Radiology. & \textbullet Lead Coordinator \newline \textbullet Competition Design \newline \textbullet Evaluation Design \newline \textbullet Advertisemen \newline \textbullet Evaluator \newline \textbullet Platform Design \newline and Engineering\\ \\

\textbf{Jinsung Yoon} is a PhD student in Electrical and Computer Engineering Department at UCLA. His main research interest has been on data imputation, model interpretation, transfer learning, and synthetic data generation using adversarial learning and reinforcement learning frameworks. He has published various papers and served as a reviewer in top-tier machine learning conferences (NeurIPS, ICML, ICLR, AAAI). & \textbullet Baseline method\newline provider \newline \textbullet Data analyzer \newline \textbullet Advise on \newline competition design  \newline \textbullet Evaluator \newline \textbullet Advertisement \\ \\

\textbf{Tavian Barnes} is a masters student studying computer systems in the David R. Cheriton School of Computer Science at the University of Waterloo.  He previously worked at Microsoft Research Montréal, primarily on NLP, dialogue systems, and reinforcement learning.  He also has experience running machine learning competitions, having helped run the TextWorld competition for the 2019 IEEE Conference on Games. & \textbullet Platform Design \newline and Engineering\newline \textbullet Evaluator\\ \\

\textbf{Paul Elbers}, MD, PhD, EDIC is a medical specialist in intensive care medicine at Amsterdam UMC, Amsterdam, The Netherlands. He also leads the Right Data Right Now research group at Amsterdam UMC that specifically aims to bring machine learning to the bedside of critically ill patients to improve their outcome. He is the deputy chair of the Data Science Section of the European Society of Intensive Care Medicine and co-chair of Amsterdam Medical Data Science, home of AmsterdamUMCdb, the first freely accessible European Intensive Care database. & \textbullet Domain expertise \newline \textbullet Data provider \newline \textbullet Advise on \newline competition design \newline \textbullet Advise on \newline evaluation design \\ \\

\textbf{Patrick Thoral}, MD, EDIC works as an intensivist, medical specialist for intensive care, at Amsterdam UMC, Amsterdam, The Netherlands. With a background of medicine as well as medical informatics, he's currently responsible for implementation of the electronic health record system in the ICU. To expedite improving patient outcomes using health care data, he played a major role in releasing AmsterdamUMCdb, the first freely accessible European Intensive Care database. & \textbullet Domain expertise \newline \textbullet Data provider \newline \textbullet Advise on \newline competition design \newline \textbullet Advise on \newline evaluation design \\ \\

\textbf{Ari Ercole} MD, PhD, FICM, FRCA, FCI is a research active intensive care attending physician at Cambridge University Hospitals NHS Foundation Trust with a PhD in physics and extensive experience in computing and ICU data modelling. He is chair of the European Society of Intensive Care Medicine Data Science Section and is a founding Fellow of the Faculty of Clinical Informatics. He has authored numerous peer-reviewed publications on the re-use of routinely ICU time-series data to improve predictions and care of intensive care patients and has been involved in a number of big-data projects such as the development of the Critical Care Health Informatics Collaborative database \cite{harris2018} and the recent DAQCORD data curation guidelines \cite{ercole2020}. & \textbullet Domain Expertise \newline \textbullet Data provider \newline \textbullet Advise on \newline competition design \newline \textbullet Advise on \newline evaluation design \\ \\

\textbf{Cheng Zhang}, PhD is a senior researcher  at Microsoft Research Cambridge, UK. She leads the Data Efficient Decision Making (Project Azua) team in Microsoft. Before joining Microsoft, she was with the statistical machine learning group of Disney Research Pittsburgh, located at Carnegie Mellon University. She is interested in both machine learning theory, including variational inference, deep generative models and sequential decision making under uncertainty, as well as various machine learning applications with social impact such as education and healthcare. She has published many papers in top machine learning venues including NeurIPS, ICML, ICLR, ICLR, UAI etc. She co-organized the Symposium on
Advances in Approximate Bayesian Inference from 2017 to 2019. & \textbullet ML Expertise \newline \textbullet Advise on \newline competition design \newline \textbullet  Advise on \newline evaluation design \newline \textbullet Advertisement \\ \\

\textbf{Danielle Belgrave}, PhD is a principal researcher at Microsoft Research Cambridge, working on the intersection of machine learning and healthcare. The primary focus of her work is on developing probabilistic models to understand personalised healthcare strategies. She has published extensively on this intersection in high impact medical journals. She is the tutorial chair of NeurIPS 2019, 2020, diversity and inclusion chair of AISTATS 2020, board member of the Deep Learning Indaba, co-organiser of the first Khipu 2019, is a board member of Women in Machine Learning, program chair of WiML 2017, and has organised several other conferences and workshops. & \textbullet ML Expertise \newline \textbullet Advise on \newline competition design \newline \textbullet  Advise on \newline evaluation design \newline \textbullet Advertisement \\ \\

\textbf{Professor van der Schaar}, PhD is John Humphrey Plummer Professor of Machine Learning, Artificial Intelligence and Medicine at the University of Cambridge, a Turing Faculty Fellow at The Alan Turing Institute in London, and Chancellor's Professor at UCLA. She is also the director of Cambridge Centre of AI in Medicine. She was elected IEEE Fellow in 2009. She has received numerous awards, including the Oon Prize on Preventative Medicine from the University of Cambridge (2018), an NSF Career Award (2004), 3 IBM Faculty Awards, the IBM Exploratory Stream Analytics Innovation Award, the Philips Make a Difference Award and several best paper awards, including the IEEE Darlington Award. She holds 35 granted USA patents.  In 2019, she was identified by National Endowment for Science, Technology and the Arts as the female researcher based in the UK with the most publications in the field of AI. & \textbullet General coordination \newline and management \newline \textbullet ML Expertise \newline \textbullet Advise on \newline competition design \newline \textbullet Advise on \newline evaluation design \newline \textbullet Advertisement \\
\end{longtable}

\subsection{Resources Provided}
Microsoft will provide 2 $\$5,000$ cash prizes for the winning team from each track. In addition to the cash prize, Microsoft have made available 50 grants of \$250 Azure cloud computing credits to students participating in the contest. Students are not required to use Azure to compete.

\bibliographystyle{plainnat}
\bibliography{refs}

\end{document}